 \documentclass[pmlr,twocolumn,10pt]{jmlr} 

\usepackage{booktabs}
\usepackage{siunitx}
\usepackage{enumitem}
\usepackage{colortbl}
\usepackage{graphics}
\usepackage{caption}
\usepackage{subcaption}
\usepackage{graphicx}
\usepackage{amsmath}
\usepackage{cleveref}
\usepackage{tikz}
\usepackage{booktabs}
\usepackage{colortbl}
\usetikzlibrary{calc, positioning}
\usepackage{xcolor}
\usepackage{multirow}
\usepackage{graphicx}
\usepackage[most]{tcolorbox}

\tcbset{
  aiboxsmall/.style={
    width=0.48\textwidth,
    top=10pt,
    colback=white,
    colframe=black,
    colbacktitle=black,
    enhanced,
    center,
    attach boxed title to top left={yshift=-0.1in,xshift=0.15in},
    boxed title style={boxrule=0pt,colframe=white,},
  }
}
\newtcolorbox{AIboxSmall}[2][]{aiboxsmall,title=#2,#1}

\definecolor{aliceblue}{rgb}{0.94, 0.97, 1.0}
\definecolor{cream}{rgb}{1.0, 0.99, 0.82}
\definecolor{eggshell}{rgb}{0.94, 0.92, 0.84}
\renewrobustcmd{\bfseries}{\fontseries{b}\selectfont}
\renewrobustcmd{\boldmath}{}
\newrobustcmd{\B}{\bfseries}

\usepackage[most]{tcolorbox}
\tcbset{
  aiboxmodern/.style={
    width=\linewidth,
    fonttitle=\bfseries,
    colbacktitle=gray!20,
    coltitle=black,
    colback=white,
    colframe=gray!50,
    left=10pt,
    right=10pt,
    top=10pt,
    bottom=10pt,
    sharp corners,
    drop shadow={black!20!white},
    enhanced,
    title style={fill=gray!20, sharp corners, drop shadow={black!25!white}},
    attach boxed title to top center={yshift=-\tcboxedtitleheight/2},
    boxed title style={boxrule=0pt,colframe=gray!50, sharp corners}
  }
}
\newtcolorbox{AIboxModern}[2][]{aiboxmodern,title=#2,#1}

\newcommand{\equal}[1]{{\hypersetup{linkcolor=black}\thanks{#1}}}

\theorembodyfont{\upshape}
\theoremheaderfont{\scshape}
\theorempostheader{:}
\theoremsep{\newline}

\jmlrvolume{225}
\jmlryear{2023}
\jmlrsubmitted{LEAVE UNSET}
\jmlrpublished{LEAVE UNSET}
\jmlrworkshop{Machine Learning for Health (ML4H) 2023} 

 \title[LLMs Accelerate Medical Annotation]{LLMs Accelerate Annotation for Medical Information Extraction} 

  \author{
  \Name{Akshay Goel}\equal{These authors contributed equally.} \Email{goelak@google.com}\\
  \Name{Almog Gueta}\footnotemark[1] \Email{almoggueta@google.com}\\
  \Name{Omry Gilon}\footnotemark[1] \Email{omrygilon@google.com}\\
  \Name{Chang Liu} \Email{changliustat@google.com}\\
  \Name{Sofia Erell} \Email{rovinsky@google.com}\\
  \Name{Lan Huong Nguyen} \Email{lanhuongnguyen@google.com}\\
  \Name{Xiaohong Hao} \Email{haoxh@google.com}\\
  \Name{Bolous Jaber} \Email{bolous@google.com}\\
  \Name{Shashir Reddy} \Email{shashir@google.com}\\
  \Name{Rupesh Kartha} \Email{rupeshkartha@google.com}\\
  \Name{Jean Steiner} \Email{jsteiner@google.com}\\
  \Name{Itay Laish} \Email{itaylaish@google.com}\\
  \Name{Amir Feder} \Email{afeder@google.com}\vspace{-2.5mm}\\
  \rule{\textwidth}{0.1pt}
  \addr Google Research
  }

\begin{document}

\maketitle

\begin{abstract}
The unstructured nature of clinical notes within electronic health records often conceals vital patient-related information, making it challenging to access or interpret. To uncover this hidden information, specialized Natural Language Processing (NLP) models are required. However, training these models necessitates large amounts of labeled data, a process that is both time-consuming and costly when relying solely on human experts for annotation. In this paper, we propose an approach that combines Large Language Models (LLMs) with human expertise to create an efficient method for generating ground truth labels for medical text annotation. By utilizing LLMs in conjunction with human annotators, we significantly reduce the human annotation burden, enabling the rapid creation of labeled datasets. We rigorously evaluate our method on a medical information extraction task, demonstrating that our approach not only substantially cuts down on human intervention but also maintains high accuracy. The results highlight the potential of using LLMs to improve the utilization of unstructured clinical data, allowing for the swift deployment of tailored NLP solutions in healthcare.
\end{abstract}

\begin{keywords}
Medical NLP, Large Language Models, Data Annotation
\end{keywords}

\section{Introduction}
\label{sec:intro}

Medical NLP holds promise for transforming healthcare by extracting valuable insights from vast amounts of unstructured clinical data \citep{feder-etal-2022-building, zhang2022section, liu2023large}. These NLP insights can improve data organization and highlight critical clinical information for users \citep{landolsi2022}. However, a significant impediment to this transformation is the challenge of obtaining high-quality labeled data for training and evaluating machine learning models specific to medical NLP tasks \citep{hakala-pyysalo-2019-biomedical}. The crux of the issue resides in the human labeling process. Manual annotation, especially when it necessitates the expertise of medical professionals, is labor-intensive and expensive. Given the intricacies of medical data, achieving a comprehensive and accurate set of text spans corresponding to specific medical entities often requires multiple annotators or iterative rounds of annotation.

\begin{figure}[t!]
\centering
\includegraphics[width=0.95\linewidth]{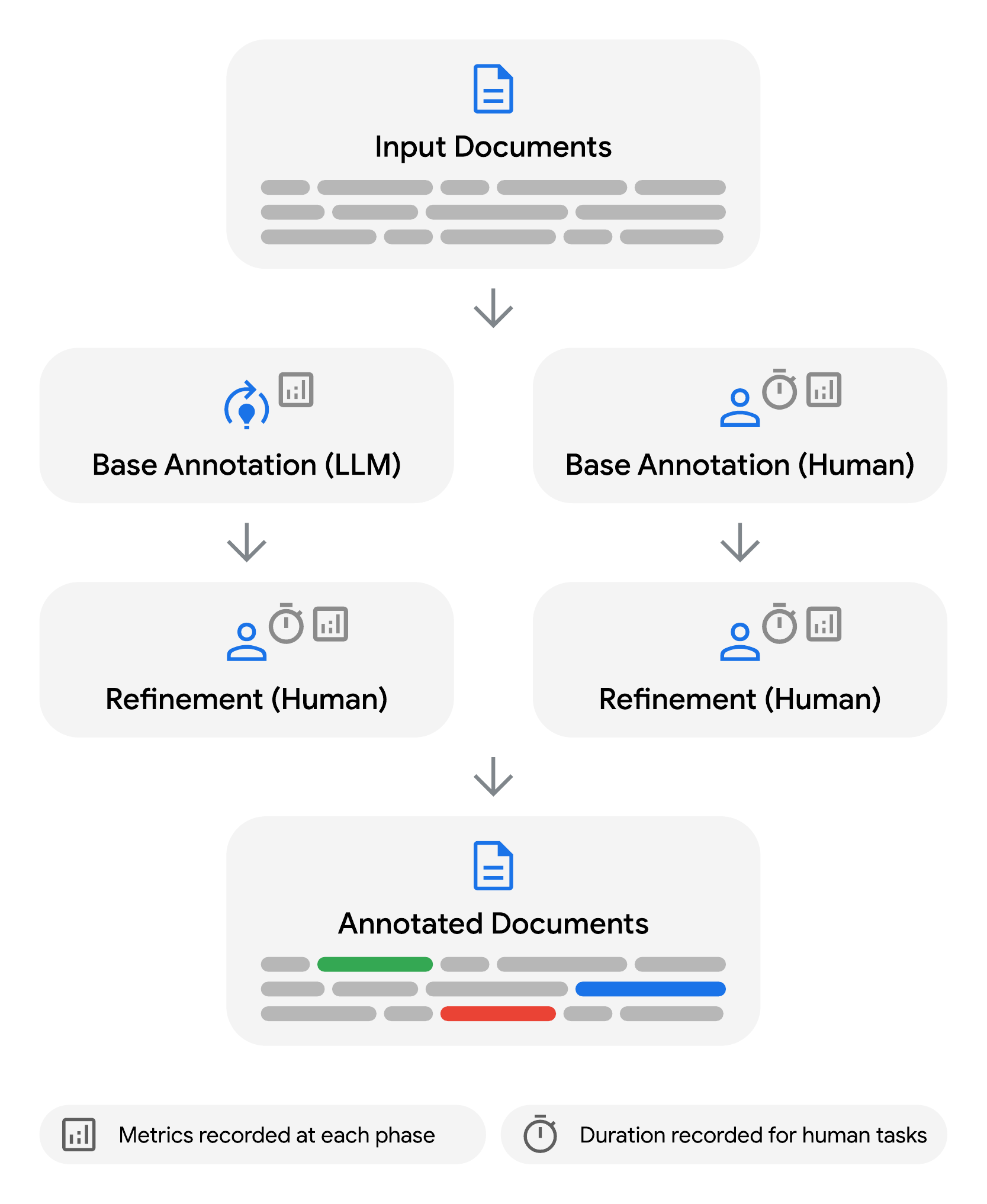}
\caption{Evaluation overview comparing annotation efficiency between LLM-assisted (left) and human-only (right) annotation workflows using the identical document set. Duration and metrics of each phase are benchmarked against \textit{i2b2} ground-truth labels.}
\vspace{-15pt}
\label{fig:project_overview_v1}
\end{figure}

To alleviate this bottleneck, in this work, we test the potential of LLMs to streamline the labeling process \citep{ding2022gpt, liu2022wanli}. LLMs have recently gained immense popularity due to their generalizability across various tasks \citep{zhou2023large}. \emph{Prompt engineering} is the optimization of natural language instructions to effectively condition an LLM \citep{reynolds2021prompt}. When incorporating examples of the task, this is termed \emph{few-shot learning} \citep{brown2020language}. Notably, \emph{prompt engineering} has shown resilience against overfitting \citep{akinwande2023}. Although LLMs offer the allure of automating data labeling, they have two main limitations: their propensity for generating unconstrained outputs and their inability to guarantee expert-level performance \citep{agrawal2022large}.

Acknowledging this, we introduce a two-step method that combines the capabilities of LLMs with human expertise. Initially, the LLM, conditioned in a \emph{few-shot learning} setup, generates \textit{Base Annotations}. Subsequently, these annotations are refined by medical annotation experts to produce \textit{Refined Annotations}.

To validate the efficacy of our approach, we compare it against a standard annotation pipeline, which utilizes a round of human Base Annotations followed by expert human Refinement Annotations. \Cref{fig:project_overview_v1} provides an overview of our evaluation, comparing an LLM-assisted annotation process to a human-only annotation process.
Our empirical evaluation, focused on the medication-extraction task, highlights the potential of our approach to significantly reduce the human annotation workload while maintaining expert-level high-quality annotations. Through this study, we aim to establish a more efficient route for labeling data to deploy robust medical NLP solutions, ultimately driving advancements in patient care and healthcare operations.

Our contributions are:
\begin{enumerate}
    \item We empirically test the utility of LLMs in annotating clinical narratives, a core bottleneck in medical NLP (see \Cref{fig:project_overview_v1}).
    \item We present a method that significantly accelerates the annotation process while maintaining expert-level quality.
  \item We produce a set of labels for medication extraction using a public medical dataset (MIMIC-IV-Note) \citep{johnson2023mimicnote}.\footnote{The labels can be accessed at PhysioNet \citep{goel2023medication, PhysioNet}).}
\end{enumerate}

The paper is structured as follows. In \Cref{sec:rel}, we discuss prior work on clinical information extraction and the challenges these systems face. In \Cref{sec:data}, we present the data and task with which we are experimenting. Subsequently, in \Cref{sec:humans} and \Cref{sec:llms}, we describe the human annotators and the LLM pipeline that supports them, respectively. Using these annotators, we then present in \Cref{sec:res} our empirical findings and discuss the efficiency gains from using our annotation approach. Finally, in \Cref{sec:disc} we discuss paths forward for utilizing LLMs to enhance medical NLP while ensuring reliability.

\section{Related Work}
\label{sec:rel}

\paragraph{Information Extraction from Clinical Notes.}
With the rise of Electronic Health Records (EHRs), there is a pressing need for efficient clinical information extraction systems \citep{Dash2019, birkhead2015uses}. 
Extracting patient information from medical notes is pivotal in clinical-NLP, with many standardized datasets and competitions \citep{uzuner2009recognizing,savova2010mayo, jensen2012mining, ford2016extracting, zhu2018clinical}. 
Modern solutions often employ the transformer architecture for these tasks \citep{vaswani2017attention, peng2019transfer, yadav2019survey, si2019enhancing, lee2020biobert}.
These approaches have shown promise in tasks like biomedical Named Entity Recognition (NER) but are constrained by their reliance on extensive domain-specific labels \citep{hakala-pyysalo-2019-biomedical}.

\paragraph{LLMs in Medical NLP.}
LLMs have transformed medical NLP, excelling in tasks such as medical question-answering and unsupervised clinical information extraction \citep{driess2023palme, singhal2022large, agrawal2022large}. Their adaptability allows for extension across various medical domains, from oncology to telemedicine, and their ability to be distilled into task-specific models, such as UniversalNER, highlights their versatility \citep{zhou2023universalner}.

\paragraph{Active Learning in Medical NLP.}
Active learning has emerged as a solution to reduce manual labeling in medical NLP. For instance, \citet{10.1093/jamia/ocv069} utilized active learning for medical concept extraction, while \citet{10.1093/jamia/ocz102} introduced a cost-aware method. \citet{FEDER2020103436} combined BERT \citep{devlin2019bert} with active learning, emphasizing iterative manual refinement. While promising, these methods require explicit model parameter tuning to fit a data distribution and task.

In this work, we propose an LLM-based approach that 
significantly enhances medical information extraction over traditional methods, without necessitating model parameter tuning.

\section{Data and Task}
\label{sec:data}

\subsection{\textit{i2b2} Dataset} 
Our experiment utilizes the $2009$ \textit{i2b2} Workshop on NLP Challenges dataset \citep{uzuner2010extracting}, consisting of de-identified clinical discharge summaries (\textit{documents}). Each document contains labels of medication \textit{fields}, which includes the medication's \textit{name}, \textit{dosage}, \textit{mode} (route) of administration, \textit{frequency}, \textit{duration} of administration, and the \textit{reason} for administration.\footnote{\citet{uzuner2010extracting} note a \textit{narrative} field, distinguishing list structure from narrative text in discharge summaries. This was omitted as our focus is classification, not extraction. Additionally, the \textit{reason} field was excluded due to its high variance in human labeling \citep{uzuner2010extracting}.} The fields of a single medication are linked together to create a \textit{medication entry}.

\subsection{Annotation Task}\label{subsec:annotation_task}
The \textit{i2b2} dataset provides high-quality labels for a medication-extraction task, divided into two sub-tasks: Named Entity Recognition (NER) for medication fields and Relationship Extraction (RE) to associate these fields with specific medication entries. This involves a two-step process: 

\begin{enumerate}
\item NER Task: Identifies the textual spans corresponding to each medication field, marking the start and end token indices.
\item RE Task: Forms medication entries by linking related fields for each medication mention.
\end{enumerate}

\begin{figure}[h]
\centering
\includegraphics[width=\linewidth]{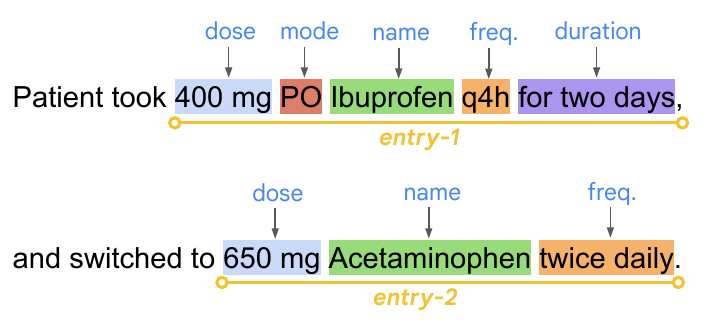}
\caption{Example of a medication-extraction annotation, showcasing two medication entries underlined in yellow. Field types are highlighted by color.}
\label{fig:annotation_example}
\end{figure}

A visual representation of the labels and task can be seen in Figure \ref{fig:annotation_example}.

\definecolor{amber}{rgb}{1.0, 0.75, 0.0}

\subsection{Data Splits \& Pre-Processing}
The dataset, comprising $261$ documents, was randomly divided into three subsets: the `pilot set' ($16$ documents, $6\%$), used to ensure human annotators' understanding of the guidelines and task and to select expert raters for the refinement task (see \S\ref{sec:humans}); the `development set' ($29$ documents, $11\%$), for selecting optimal prompts and parameters for the LLM pipeline; and a `test set' ($216$ documents, $83\%$), for measuring the performance of both annotation methods, with human or LLM Base Annotations. Table \ref{tab:field_counts} presents the number of occurrences of each field in each dataset subset.

\begin{table}[h]
\centering
\resizebox{\linewidth}{!}{%
\begin{tabular}{l|ccccc}
\toprule
  \textbf{} & \textbf{Name} & \textbf{Dose} & \textbf{Mode} & \textbf{Frequency} & \textbf{Duration} \\
\midrule
Total &  $9,237$ &  $4,646$ &  $3,491$ & $4,182$ & $562$ \\
Pilot   & $511$   & $300$   & $180$  & $282$  & $38$  \\
Development   &  $1,123$ &   $523$ &   $437$ & $452$ & $57$ \\
Test  &  $7,603$ &  $3,823$ &  $2,874$ & $3,448$ & $467$ \\
\bottomrule
\end{tabular}%
}
\caption{Number of occurrences for each labeled field across all data subsets of the \textit{i2b2} 2009 dataset.}
\label{tab:field_counts}
\end{table}

Minimal preprocessing was performed on the dataset to remove leading and trailing punctuation, spaces, and tabs from label spans. In addition, discontinuous medication fields were infrequent, constituting less than 1\% of the data, and were therefore excluded to streamline the preprocessing stage.

\section{Human Annotations}
\label{sec:humans}

Human annotations in the medical domain often necessitate medical expertise and extensive training, making the process both time-consuming and costly.  Annotation typically involves marking spans of text in lengthy documents, requiring high focus to avoid errors. Rater fatigue is thus a significant concern.

\paragraph{Annotation Workflow:}
To mitigate the above challenges, our annotation process comprises two sequential phases: 
\begin{enumerate}
    \item \textbf{Base Annotation Phase}: Initial labels are added to blank documents by a rater.
    \item \textbf{Refinement Annotation Phase}: The base labels are refined and corrected by specialized \emph{expert raters}.
\end{enumerate}

\paragraph{Annotation Platform and Guidelines}
Annotations were carried out on a text labeling platform, configured to support the medication-extraction schema detailed in \S\ref{subsec:annotation_task}. The platform consists of a viewing panel to display raw medical documents and a question panel to guide the annotation process. Special cases like ambiguity between frequency and duration were handled according to specific rules based on the original \emph{i2b2} guidelines.

\paragraph{Raters Team Composition and Expertise}
Our annotation team was composed of seven raters, each with a medical background and an extensive experience in medical NLP tasks. Their medical training ranged from medical assistant to attending physician.

A pilot task was conducted to select the \emph{expert raters} based on their performance in annotating a pilot set of documents, resulting in the selection of three expert raters. To optimize the use of rater time and expertise, the more exact Refinement Annotations were performed solely by \emph{expert raters}. Additionally, both \emph{expert raters} and \emph{non-expert raters} contributed to Base Annotations. This setup allowed us to perform a subgroup analysis on expertise-level for the annotation pipeline.

\paragraph{Rater Tasks}
Raters were responsible for three types of tasks, where documents were randomly allocated to ensure no rater annotated a document twice:
\begin{enumerate}
    \item \textbf{Base Rater Label}: Base Annotation of medical documents (labeling documents from scratch). Performed by \emph{all raters}.
    \item \textbf{Refined Rater Label}: Refinement Annotation of labeled documents produced by another rater. Performed by \emph{expert raters}.
    \item \textbf{Refined LLM Label}: Refinement Annotation of labeled documents produced by the LLM pipeline. Performed by \emph{expert raters}.
\end{enumerate}

\section{LLM Annotations}
\label{sec:llms}

To assess the efficiency gains from LLM annotation in medication extraction, we devised an LLM labeling pipeline that consists of three key components: 
(1) an LLM, 
(2) a task-specific prompt with an input chunk of text, and 
(3) a post-processing \textit{Resolver Module} that converts the generative outputs into NER-RE structured objects.

The steps for the LLM medication-extraction annotation pipeline are:

\begin{enumerate}
\item Input document is segmented into text chunks, which are delineated by textual breaks such as new lines and end-of-sentence markers.
\item Each text chunk is incorporated into a task-specific \emph{few-shot prompt} designed for medication extraction.
\item The LLM performs inference on this prompt, generating textual outputs.
\item Generative textual outputs are converted into structured medication-extraction objects by the resolver module. Each object encapsulates the relevant medication fields along with their associated RE information.
\item Optionally, human reviewers can refine these annotations to enhance their quality.
\end{enumerate}

\begin{figure}[t!]
\centering
\includegraphics[width=0.90\linewidth]{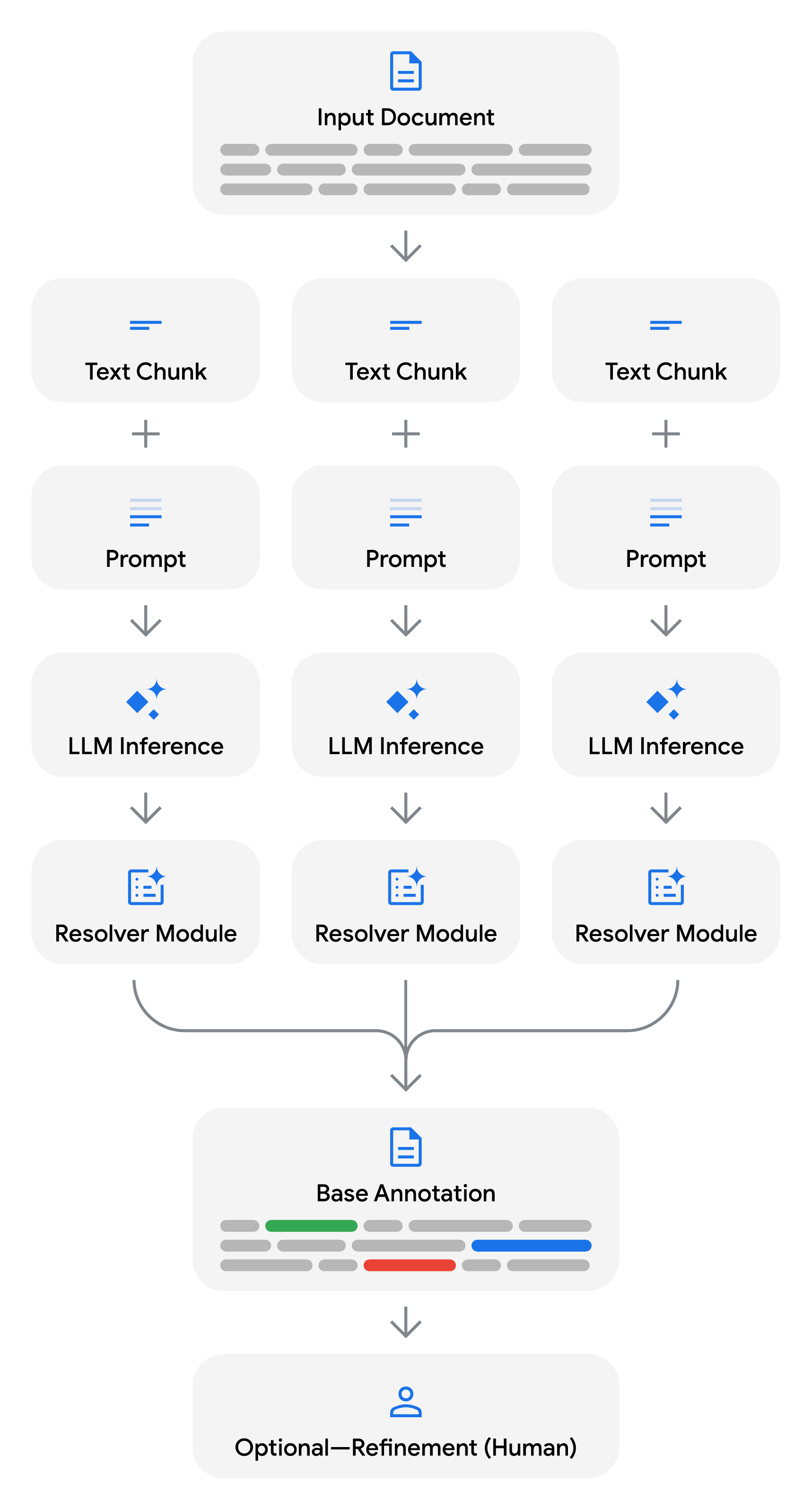}
\caption{Overview of the LLM medication-extraction annotation pipeline. Our labeling pipeline consists of a task-specific prompt with an input chunk of text, an LLM, and a post-processing Resolver Module that converts the generative outputs into NER-RE structured objects.}
\vspace{-15pt}
\label{fig:llm_pipeline}
\end{figure}

\Cref{fig:llm_pipeline} provides an illustration of the pipeline. 

\paragraph{PaLM 2 Language Model.}
To construct the LLM medical-extraction annotation pipeline, we utilized Google's latest-generation language model, PaLM 2 \citep{anil2023palm}. This model has achieved state-of-the-art performance across a broad spectrum of tasks. While employing a fine-tuned version of PaLM 2 for the NER-RE task might yield even better results, our primary objective was to demonstrate a proof-of-concept using an LLM, without explicit tuning.

\subsection{Prompt Structure}
\begin{figure}[!ht]
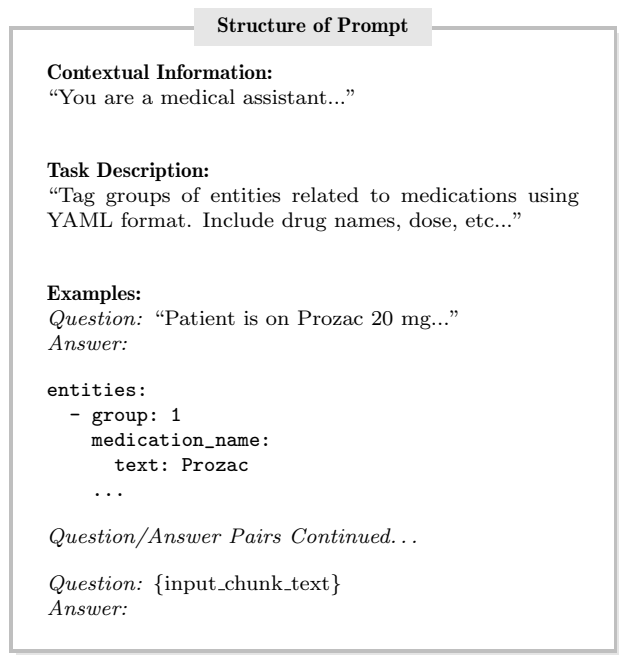

    \begin{AIboxModern}{\footnotesize Structure of Prompt}
    \footnotesize
    \textbf{Contextual Information:}\\
    ``You are a medical assistant..." \\[1em]

    \textbf{Task Description:}\\
    ``Tag groups of entities related to medications using YAML format. Include drug names, dose, etc..." \\[1em]

    \textbf{Examples:}\\
    \textit{Question:} ``Patient is on Prozac 20 mg..." \\
    \textit{Answer:}
    \begin{center}
    \begin{verbatim}
entities:
  - group: 1
    medication_name:
      text: Prozac
    ...\end{verbatim}
    \end{center}
    \textit{Question/Answer Pairs Continued\ldots} \vspace{1em} \\
    \textit{Question:} \{input\_chunk\_text\}\\
    \textit{Answer:}
    \end{AIboxModern}
    \caption{Prompt structure for LLM annotation: includes contextual information, task description, and curated examples to illustrate task intricacies.}
    \label{fig:prompt_structure}    
\end{figure}

To apply a generalized LLM to the specific medication-extraction task, we utilize \emph{in-context} or \emph{few-shot learning} \citep{brown2020language}. This approach leverages prompts containing: contextual information, a description of the task, and one or more \emph{few-shot examples} with input-output pairs. During inference, the input chunk text is integrated into the prompt, as illustrated in Figure \ref{fig:prompt_structure}. (See \cref{app:promptexample} for example prompt implementations).

\subsection{Medication-Extraction Prompts Schema}

To leverage in-context learning for medication extraction, the task should be precisely described and defined within the prompt. This information shapes the generative outputs from the LLM. We evaluated two prompt schemas, and generated annotations of both of which were ultimately ensembled to achieve optimal performance. The first schema, \textit{IOB-Token}, utilizes a modified version of the Inside-Outside-Beginning (IOB) token-tagging technique \citep{ramshaw1999text}, using medication fields as possible classes. The IOB method classifies each token into a specific class. Our modification introduces an additional group tag to retain RE information, as depicted in Figure \ref{fig:iob_token}.

\begin{figure}[h]
    \begin{AIboxModern}{\footnotesize IOB-Token Annotation}
    \footnotesize
        \textit{Chunk Input:} ``Patient takes Prozac 20 mg and Zoloft 50 mg...''

        \vspace{1em} 
        
        \begin{center}
            \begin{tabular}{l l l}
                \textbf{Token} & \textbf{IOB Tag-Label} & \textbf{Group Tag} \\
                \hline
                `Patient' & O & None \\
                `takes' & O & None \\
                `Prozac' & B-MEDICATION & `entity\_1' \\
                `20' & B-DOSE & `entity\_1' \\
                `mg' & I-DOSE & `entity\_1' \\
                `and' & O & None \\
                `Zoloft' & B-MEDICATION & `entity\_2' \\
                `50' & B-DOSE & `entity\_2' \\
                `mg' & I-DOSE & `entity\_2' \\
            \end{tabular}
        \end{center}
        
    \end{AIboxModern}
    \caption{Example of IOB-token annotations for medication information extraction. In the prompt implementation, the tokenized text, IOB tag, and group entity are comma-delimited.} 
    \label{fig:iob_token}    
\end{figure}

The second schema, \textit{Direct Chunk}, directly produces the fields labels and grouping for the input chunk of text which may contain multiple words (e.g., `20 mg'), without requiring intermediate tokens. Figure \ref{fig:direct_chunk} provides an illustration of this schema.

\begin{figure}[h]
    \begin{AIboxModern}{\footnotesize Direct Chunk Annotation}
    \footnotesize
        \textit{Chunk Input:} ``Patient takes Prozac 20 mg and Zoloft 50 mg...''

        
        \begin{center}
        \resizebox{\linewidth}{!}{%
            \begin{tabular}{l l l}
                \textbf{Field Type} & \textbf{Entity Text} & \textbf{Position (start-end)} \\
                \hline
                \textbf{Group: 1} & & \\
                MEDICATION: & Prozac & 18-24 \\
                DOSE: & 20 mg & 25-30 \\
                FREQUENCY: & - & \\
                DURATION: & - & \\
                REASON: & - & \\
                MODE: & - & \\
                \hline
                \textbf{Group: 2} & & \\
                MEDICATION: & Zoloft & 35-41 \\
                DOSE: & 50 mg & 42-47 \\
                FREQUENCY: & - & \\
                DURATION: & - & \\
                REASON: & - & \\
                MODE: & - & \\
            \end{tabular}%
        }
        \end{center}
        
    \end{AIboxModern}
    \caption{Example of Direct Chunk annotations, demonstrating the extraction of medication fields from the input by groups. In the prompt implementation, each group is represented in YAML format.}
    \label{fig:direct_chunk}        
\end{figure}

\newpage
Once generative outputs are produced, the \textit{Resolver Module} transforms unstructured output data into medication-extraction objects. The object definitions of the output are consistent with the human labeling pipeline, facilitating seamless integration between LLM and human annotations. Both the IOB-Token and Direct Chunk schemas have a dedicated resolver module. For \textit{IOB-Token resolver}, lines are parsed using a comma as the delimiter, while the \textit{Direct Chunk resolver} parses output using a Python YAML library. See \cref{app:resolver} for further details.

\subsection{Prompt Engineering \& Ensembling}\label{sec:prompt_engineering}

With an established prompt structure and schema in place, we performed iterative experiments to identify the optimal LLM prompts. To identify areas for prompt improvement, we evaluated errors in the `development set'. The most significant error patterns were corrected and incorporated as instructions and examples into subsequent prompt versions (see \cref{app:erroreval} for further details). We ultimately developed 10 versions of the IOB-Token schema prompt and 5 versions of the Direct Chunk prompt. See \cref{app:llm_inference_time} for details on LLM document inference time.

We also evaluated other key parameters: Our early observations indicated that a temperature of \(0.0\) offered the best performance. Max-decoding-steps above \(1024\) tokens had minimal impact, so this value was fixed. Chunk sizes of $250$ and $180$ offered the best performance for IOB-Token and Direct Chunk, respectively. See \cref{app:hyperparam} for further discussion.

We placed a higher priority on annotation recall over precision because our empirical experience suggests that deleting annotations is less time consuming than producing them from scratch. To select for a weighted balance of recall over precision, we used an F2 score.
\Cref{app:typesrefinment} quantifies the benefit of prioritizing recall over precision, showing the lower human time cost for deleting a label compared to adding or modifying one. 

Lastly, we evaluated the performance of an ensemble that combines generated annotations from the best configurations from both prompt schemas. Pseudocode for the ensemble procedure is presented in \cref{app:ensemble_code}. The ensemble had the highest F2 score and was therefore chosen for the LLM Base Annotations on the `test set'. See \cref{app:final_model} for experiment results of best performing configurations.

\newpage
\section{Results}
\label{sec:res}

We focus on three key aspects: time savings, label quality, and the performance differentiation between expert and non-expert raters for the Base Annotations phase. Our main findings are as follows:

\begin{itemize}
    \item \textbf{Time Efficiency:} 
    Using the LLM pipeline to replace the human rater in the initial Base Annotations phase reduces human annotation total time by 58\% on average.
    
    \item \textbf{Label Quality:} 
    Comparative evaluations between human raters and the LLM pipeline show that the label quality at the Base Annotations phase is comparable. Additionally, the expert Refinement phase achieves the same quality level regardless of the source of Base Annotations — whether from human raters or the LLM pipeline.
    
    \item \textbf{Expert vs. Non-Expert Performance:}
    Subgroup analysis focusing on \emph{expert raters} demonstrates that LLMs improve Base Annotations efficiency by 26\% on average, underlining the utility of LLMs even within exceptionally skilled raters. Furthermore, the benefit of utilizing LLMs is even more pronounced compared to \emph{non-expert raters}.
\end{itemize}

In addition, to substantiate these findings, we have employed statistical tests for the main experiments. The details of which can be found in Table~\ref{tab:comparative_significance}.

\begin{table*}[ht]
\centering
  \begin{tabular}{l|ccc|ccc|cc}
    \multicolumn{1}{c|}{\textbf{Label Type}} &
      \multicolumn{3}{c|}{\textbf{Vertical}} &
      \multicolumn{3}{c|}{\textbf{Horizontal}} &
      \multicolumn{2}{c}{\textbf{Time Cost}} \\
    & Recall & Precision & F1 & Recall & Precision & F1 & Mean & Median  \\
    \midrule
    Base Rater & 0.789 & 0.893 & 0.838 & 0.734 & 0.821 & 0.775 & 17.60 & 11.93 \\
    Base LLM & 0.850 & 0.762 & 0.804 & 0.810 & 0.703 & 0.752 & n/a & n/a \\
     Base Rater + Refinement &  0.912 &  0.907 &  0.910 &  0.887 &  0.879 &  0.883&  26.67 &  19.18 \\
     Base LLM + Refinement &  0.921 &  0.893 &  0.907 &  0.892 &  0.860 &  0.876&  11.32 &  7.27 \\
    \bottomrule
  \end{tabular}
  \caption{Labeling Quality and Time metrics for the entire Test Set (n=216 documents). Quality metrics are calculated at the phrase-level. \emph{Time Cost} refers to the total time expended by raters, measured in minutes per document. Refinement methods demonstrate significantly higher quality, with LLM + Refinement method reducing annotation time by $57.6\%$, while compromising only $1\%$ in quality metrics.
  }
 \label{tab:overall_metrics}
\end{table*}

\begin{figure}[ht]
\centering
\includegraphics[width=\linewidth]{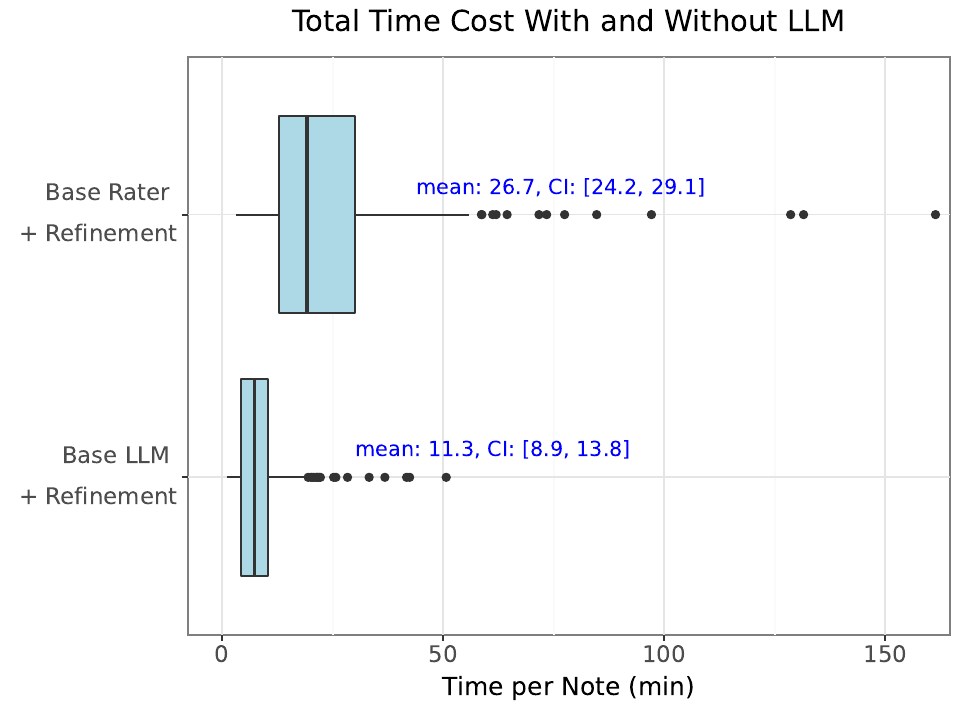}
\caption{Total human time cost for producing refined labels with and without LLM. Using LLMs reduces $58\%$ of the annotation time ($11.3$ vs. $26.7$).}
\label{fig:time_cost_overall}
\end{figure}

\begin{figure}[ht]
\centering
\includegraphics[width=\linewidth]{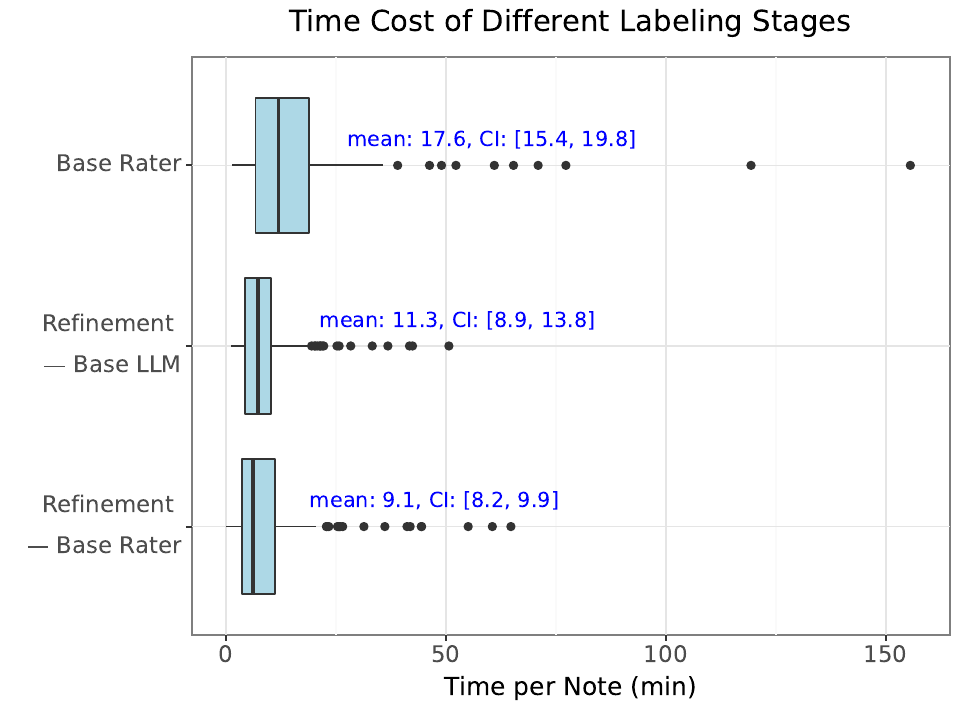}
\caption{Human time cost at the two labeling phases- Base Annotations and Refinement. Refining annotations reduces time costs by $42\%$ on average compared to starting from scratch.}
\label{fig:time_cost_stages}
\end{figure}

\subsection{Efficiency Metrics.}
Figure~\ref{fig:time_cost_overall} and Table~\ref{tab:overall_metrics} illustrate that the integration of LLMs leads to a substantial decrease in the human time cost for the production of \emph{expert-level} labels. Using LLMs, the mean time required to produce refined labels for a document decreased by 58\%, from \(26.67\) minutes to \(11.32\) minutes (\(p < 1 \times 10^{-18}\)). Figure~\ref{fig:time_cost_stages} provides a more granular view of the time costs of each annotation phase. The \emph{Base Rater} has the highest time cost, with a mean of \(17.6\) minutes per document, compared to the \emph{Refinement over Base LLM} and \emph{Refinement over Base Rater}, which have means of \(11.3\) and \(9.1\) minutes per document, respectively.

\subsection{Quality Metrics.}\label{sec:quality_metrics}
To evaluate label quality, we utilize phrase-level evaluation\footnote{Token-level evaluation shows similar results and is presented in Appendix~\ref{app:token_level_evaluation}.} based on vertical and horizontal metrics defined by ~\citep{uzuner2010extracting}. \emph{Phrase-level} refers to the entire text span of a field, as opposed to \emph{Token-level} which refers to each token in the span, e.g. `20 mg' vs. [`20', `mg'].  \emph{Vertical metrics} evaluate NER with respect to fields, while \emph{horizontal metrics} also evaluate RE between fields and the medication entry. Pseudocode for these metrics is presented in \cref{app:eval_metrics_code}. Table~\ref{tab:overall_metrics} outlines the label quality across the two phases of labeling, with and without the LLM. 

\paragraph{Quality of Base Annotations Phase.}
Comparing the Quality of \emph{Base Rater Labels} and \emph{Base LLM Labels} in Table~\ref{tab:overall_metrics} indicates that the {Base rater} outperforms in F1 score (\(0.838\) vs. \(0.804\) in vertical F1 score, and \(0.775\) vs. \(0.752\) in horizontal F1 score), due to the dominance in precision difference, while the \emph{Base LLM} obtained a higher recall rates, as we intended (see Sec~\ref{sec:prompt_engineering}). 

\begin{table*}[ht]
\centering
\resizebox{\textwidth}{!}{%
\begin{tabular}{l|ccc|ccc|cc}
    \multicolumn{1}{c|}{\textbf{Label Type}} &
    \multicolumn{3}{c|}{\textbf{Vertical}} &
    \multicolumn{3}{c|}{\textbf{Horizontal}} &
    \multicolumn{2}{c}{\textbf{Time Cost}} \\
    & Recall & Precision & F1 & Recall & Precision & F1 & Mean & Median  \\
    \midrule
    Base Expert Rater &  0.908 &  0.893 &  0.900 &  0.882 &  0.859 &  0.871 &  16.17 &  11.13 \\
    Base Expert Rater + Refinement & 0.923 & 0.900 & 0.911 & 0.898 & 0.873 & 0.886 & 22.25 & 15.22 \\
    Base LLM & 0.847 & 0.762 & 0.802 & 0.815 & 0.706 & 0.757 & n/a & n/a \\
    Base LLM + Refinement &  0.914 &  0.885 &  0.899 &  0.896 &  0.858 &  0.876 &  11.97 &  7.55 \\
    \cline{1-9}
    \midrule
    Base Non-Expert Rater &  0.712 &  0.892 &  0.792 &  0.637 &  0.789 &  0.705 &  18.67 &  12.36 \\
    Base Non-Expert Rater + Refinement & 0.905 & 0.913 & 0.909 & 0.880 & 0.882 & 0.881 & 30.41 &  21.06 \\
    Base LLM &  0.853 &  0.763 &  0.805 &  0.806 &  0.700 &  0.750 &  n/a &  n/a \\
   Base LLM + Refinement & 0.925 & 0.899 & 0.912 & 0.890 & 0.861 & 0.875 & 10.83 & 7.10 \\
    \bottomrule
\end{tabular}%
}

\caption{Comparison of Quality and Time metrics in Base Annotations by Experts and Non-Experts. \textbf{Top:} Test documents with Base Annotations performed by experts (n=93). \textbf{Bottom:} Test documents with Base Annotations performed by non-experts (n=123). Base LLM and Base LLM + Refinement are included for completeness. Note: Refinements are exclusively expert-performed. \emph{Time Cost} denotes the total time spent by raters, in minutes per document. 
Notably, the quality of Base annotations by experts without refinement is comparable to that of Base LLM + Refinement, but requires 26\% more time.}
\label{tab:subgroup_metrics}
\end{table*}

\begin{table*}[ht]
\centering
\resizebox{\textwidth}{!}{%
  \begin{tabular}{l|ccc|ccc|cc}
    \multicolumn{1}{c|}{\textbf{Label Comparison}} &
      \multicolumn{3}{c|}{\textbf{Vertical}} &
      \multicolumn{3}{c|}{\textbf{Horizontal}} &
      \multicolumn{2}{c}{\textbf{Time Cost}} \\
    & $\Delta_{\textbf{Recall}}$ & $\Delta_{\textbf{Precision}}$ & $\Delta_{\textbf{F1}}$ & $\Delta_{Recall}$ & $\Delta_{\textbf{Precision}}$ & $\Delta_{\textbf{F1}}$ & $\Delta_{\textbf{Median}}$  \\
    \midrule
    Base LLM+Refin. $-$ Base Rater+Refin. & $0.009^{**}$ & $-0.014^{**}$ & $-0.003$ & $0.006$ & $-0.018^{**}$ & $-0.007^{*}$ & $-11.90^{**}$ \\
    Base LLM+Refin. $-$ Base Expert Rater & $0.006^{**}$ & $-0.008^{**}$ & $-0.001$ & $0.013^{**}$ & $-0.001$ & $0.006^{*}$ & $-3.58^{**}$  \\
    Base LLM \qquad$-$ Base Non-expert Rater & $0.141^{**}$ & $-0.129^{**}$ & $0.013^{**}$ & $0.169^{**}$ & $-0.089^{**}$ & $0.045^{**}$ & $-12.36^{**}$ \\
    \bottomrule
  \end{tabular}%
  }
\caption{Comparing Quality and Time metrics across labeling methods on the Test Set. `Refin.' denotes Refinement. Differences ($\Delta$) are calculated by subtracting the second method's values from the first. ** and * signify statistical significance at $p<0.01$ and $p<0.05$. See \Cref{app:stats} for analysis details.}
   \label{tab:comparative_significance}
\end{table*}

\paragraph{Quality of Refinement Annotations Phase.}
Upon transitioning to the Refinement phase, a striking improvement in label quality is evident from the results presented in Table~\ref{tab:overall_metrics}. Both \emph{Base LLM + Refinement} and \emph{Base Rater + Refinement} exhibit substantial advancements in their F1 scores. Specifically, the F1 score for \emph{Base LLM + Refinement} reaches \(0.907\) in vertical metrics and \(0.876\) in horizontal metrics. These scores closely mirror the high quality of the \emph{Base Rater + Refinement}, which achieves F1 scores of \(0.910\) and \(0.883\) for vertical and horizontal metrics, respectively. An analysis of the precision and recall metrics demonstrates similar alignment.

Results suggest a noteworthy convergence in label quality between the \emph{Base LLM + Refinement} Labels and \emph{Base Rater + Refinement} Labels. The findings also highlight the value of expert rater involvement in elevating both LLM-generated and average-rater annotations to produce high-quality labels.

\subsection{Impact of Rater Expertise}
To further evaluate the quality and efficiency metrics between expert raters and non-expert raters, we evaluated two subgroups of test documents, differing based on the rater's skill level for the Base Annotation phase.
 
 Table~\ref{tab:subgroup_metrics} shows that the time cost for producing Base Annotations is similar regardless of the level of expertise. The mean time is 16.17 minutes for experts and 18.67 minutes for non-experts.\footnote{A Figure displaying the time cost for experts versus non-experts, as well as the time cost per rater, are shown in \cref{app:time_cost_raters}.} 
 
 Evaluating quality, Table~\ref{tab:subgroup_metrics} also underscores the strong relationship between rater expertise and labeling quality. \emph{Base expert raters} labels achieve high F1 scores—\(0.900\) in vertical and \(0.871\) in horizontal metrics—signifying that these labels seldom require refinement. However, the efficiency aspect should not be overlooked. The table demonstrates that even \emph{expert raters} realize significant time savings when assisted by LLMs. Specifically, the average time for labeling reduces from \(16.17\) minutes with \emph{Base Expert Rater Labels} to \(11.97\) minutes with \emph{Base LLM + Refinement Labels}, improving efficiency by 26\% (\(p < 0.01\)). Lastly, for the \emph{non-expert raters} subgroup, the benefits of using LLMs are even more pronounced, as can be seen in Table~\ref{tab:subgroup_metrics}. 

\section{Discussion}
\label{sec:disc}

Our study demonstrates that LLMs can significantly accelerate the process of medication information extraction, achieving baseline accuracy comparable to that of a trained medical NLP annotator, with superior recall at the expense of precision. Remarkably, the results were achieved solely through prompt engineering, without direct model parameter tuning. Furthermore, we show that when integrated into a human-in-the-loop process, LLMs can facilitate the generation of expert-level annotations while saving considerable human time. This is particularly noteworthy given the traditionally high costs and time investments required for generating high-quality medical NLP labels through human annotation.

In future work, we plan to explore fine-tuning LLMs for specific tasks to improve performance, and the integration of constrained decoding to ensure outputs adhere to predefined schemas. As the field continues to advance rapidly, we expect to see further improvements in performance and utility.

In addition to speeding up the annotation process, our study reveals the importance of expert-level human annotators in achieving high-quality results. We quantified the time-efficiency gains from using LLMs in the annotation pipeline and found that optimizing for recall is more time-efficient as deleting annotations is easier than adding new annotations. Although our study focused on medical information extraction, the flexibility of LLMs suggests their applicability across various information extraction tasks.

However, our study is not without its limitations. The focus on a single domain, namely medication extraction, may limit the generalizability of our findings. As more diverse and validated medical NLP datasets become available, future studies could broaden this scope. While our study concentrated on the feasibility of LLM-based labeling, a key area for further investigation involves directly comparing an LLM annotation pipeline with other ML methods, including active learning. Lastly, our cost analysis solely considers a basic metric of human time cost. It is important to differentiate between expert-level and average raters in future assessments, and to explore a cost-aware approach.

We believe that integrating LLMs into data labeling workflows has potential to change the field, especially in areas where high-quality annotations are required but resources are limited. As LLMs continue to evolve, they are poised to make the challenge of data acquisition increasingly manageable.

\acks{The authors extend their sincere gratitude to Dem Gerolemou for substantial contributions to the visuals, and to Wei-Hung Weng for offering insightful feedback during the manuscript's review process.}
\clearpage
\bibliography{goel23}

\clearpage
\appendix

\appendix
\section{LLM Labeling Pipeline Implementation Details}

Here we describe implementation details for our LLM pipepline, including our resolver module, configurations and hyper-parameters and error identification and improvement.

\subsection{Resolver Module: Parsing Generative Outputs from Unconstrained LLMs}
\label{app:resolver}
Parsing generative outputs from an unconstrained LLM can present various edge cases that are specific to the annotation schema. In the case of the IOB-Token schema, the outputs from the LLMs were formatted so that each token was separated by a new line. As such, the outputs were processed one line at a time. The individual components of each token were then parsed using a comma-delimited structure and regular expressions. Occasionally, the LLM produced outputs that did not conform to the expected structure, such as an extra comma or a missing new line. When the output structure could not be resolved, it was discarded and logged for further analysis. The `Group Tag' information is employed to group fields into medication entries.

Similarly, for the Direct-Chunk implementation, the LLM output is produced in YAML format and delimited by backticks (\texttt{```}). This output is isolated and then sent to the standard Python YAML parsing library to generate a dictionary representation of the output. If an edge case leads to a YAML parsing error, the output is logged for further analysis. Once the output has been parsed into a dictionary, each entity is mapped back to the original text. Since the fields for a specific medication are all contained within a single YAML output, individual grouping of fields is unnecessary.

During the phase where outputs are matched to the original text, additional considerations come into play, such as applying a ``fuzzy match." This allows for some minor differences in characters and helps to improve entity recall (i.e., matching a correctly spelled medication generated by the LLM with a misspelled medication in the raw text). When the outputs could not be resolved to fit the target structure or mapped back to the original text, they were logged for analysis but ultimately not used as labels. This can contribute to potential recall errors.

\subsection{Error Identification and Prompt Improvement}
\label{app:erroreval}
Our experiments with the LLM pipeline revealed various types of errors, which have been categorized and addressed to improve in-context learning. Errors spanned from missing annotations to spurious annotations and field-type confusions. 

\textbf{Missing Annotations:} The model had difficulty in identifying less frequent or differently formatted fields. We incorporated additional examples to address these omissions, though it led to a trade-off in overall precision.

\textbf{Spurious Annotations:} The model sometimes incorrectly annotated entities that did not belong to medication information. To mitigate this, we added negative examples in the prompt to refine its identification capabilities.

\textbf{Confusion between Fields:} Occasionally, the model confused similar field types. This was particularly evident in distinguishing between `mode' and `frequency'. Adjustments were made in the prompt to improve the model's accuracy in this aspect.

\textbf{Nuanced Errors:} Many errors exhibited a reasonable basis, pointing to the complexities in the common definitions used in the medical domain. For example, the model classifying `oxygen' as a medication could be attributed to the overlap between the common definitions of chemical medications and therapeutic treatments like oxygen therapy. Examples of these nuanced errors were also incorporated into the prompt.

\subsection{Configurations and Hyper-Parameters}
\label{app:hyperparam}
In addition to experimenting with different prompts, we also examined various hyper-parameters, including chunk size, maximum decoding steps, and temperature. We tested multiple values for each parameter and selected those that performed best. It is worth noting that we conducted a partial grid search and drew our conclusions based on these results. To determine the best configuration we used a balanced F2 score defined as \( F_2 = \frac{5}{{\frac{4}{\text{{recall}}} + \frac{1}{\text{{precision}}}}} \). This provided a higher emphasis on recall, which we believed would be more important for the annotation task.
For all best-performing models, 1024 maximum decoding steps yielded the best performance. The optimal chunk size varied between YAML-based and IOB-based models, with chunks of 180 and 250 characters performing best, respectively. Different optimal parameters between the two schema were expected given their different approaches. The YAML-based models directly produce labels and groups for the input chunk which may benefit from shorter chunks, whereas IOB-based models have an intermediate step of outputting per token before grouping them. For all selected models, a temperature of 0.0 proved optimal, supporting that a deterministic setting is optimal for an information extraction task.

\begin{table*}[!ht]
  \centering
  \small
  \begin{tabular}{lcccccc}
  \toprule
  \textbf{Type} & \textbf{Chunk Size} & \textbf{F1 Score} & \textbf{F2 Score} & \textbf{Recall} & \textbf{Precision} \\
  \midrule
  IOB Token                 & 250   &  \textbf{82.0} & 84.0          & 85.5   & 78.7 \\
  Direct Chunk              & 180         & 80.9          & 80.8          & 80.7   & \textbf{81.0} \\
  \rowcolor{gray!15}Ensemble & -           & 81.8          & \textbf{84.9} & \textbf{87.1}   & 77.1\\
  \bottomrule
  \end{tabular}
  \captionsetup{width=\linewidth}
  \caption{Performance of top experiments on development set for both prompt schemas. F2 score was used to select top experiments. The ensemble version, highlighted in gray, was the final version used for LLM base annotations on the test set.}
  \label{tab:models_performance}
  \end{table*}

\subsection{Pseudocode for the Ensemble Procedure to Combine Annotations}\label{app:ensemble_code}
\begin{figure}[h!]
    \centering
    \includegraphics[width=0.5\textwidth]{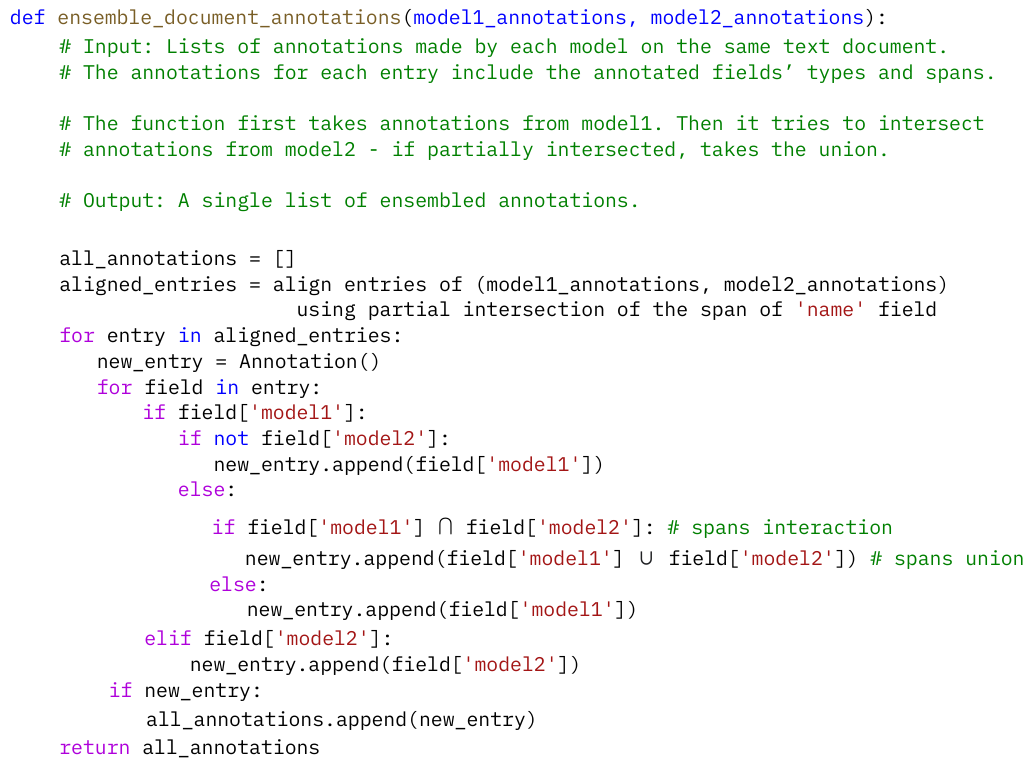} 
    \caption{Pseudocode for the ensemble procedure.}
    \label{fig:ensemble_code}
\end{figure}

Figure \ref{fig:ensemble_code} presents pseudocode for the ensemble procedure that combines generated annotations from both prompt schemas. As described in Section \ref{sec:prompt_engineering}, the ensemble achieved the highest F2 score and was therefore chosen for the LLM Base Annotations on the `test set'.
  
\subsection{Choosing the Final Model for LLM Base Annotations}\label{app:final_model}
Table \ref{tab:models_performance} displays the performance of the best-performing models on the development set. We observe that the ensemble of the two top models achieves the best F2 score, indicating the best balance of precision and recall for our objective. Consequently, we selected the ensemble model for the LLM Base Annotations.

\section{Inference Time of LLM Annotations}\label{app:llm_inference_time}
As the inference process can be scaled across multiple documents simultaneously with appropriate infrastructure, a direct comparison of LLM inference time to human annotation time is not applicable. Nonetheless, performing inference using PaLM 2 on a TPU 8x8 pod, applied individually to each document in a random subset comprising 15\% of the documents, revealed an average inference time of 1.5 minutes per 1,000 characters, with a standard deviation of 0.77 minutes. This results in an average of 9.49 minutes per document, with a standard deviation of 5.55 minutes.
  
\section{Pseudocode for Vertical and Horizontal Metrics}\label{app:eval_metrics_code}
\begin{figure}[h!]
    \centering
    \includegraphics[width=0.5\textwidth]{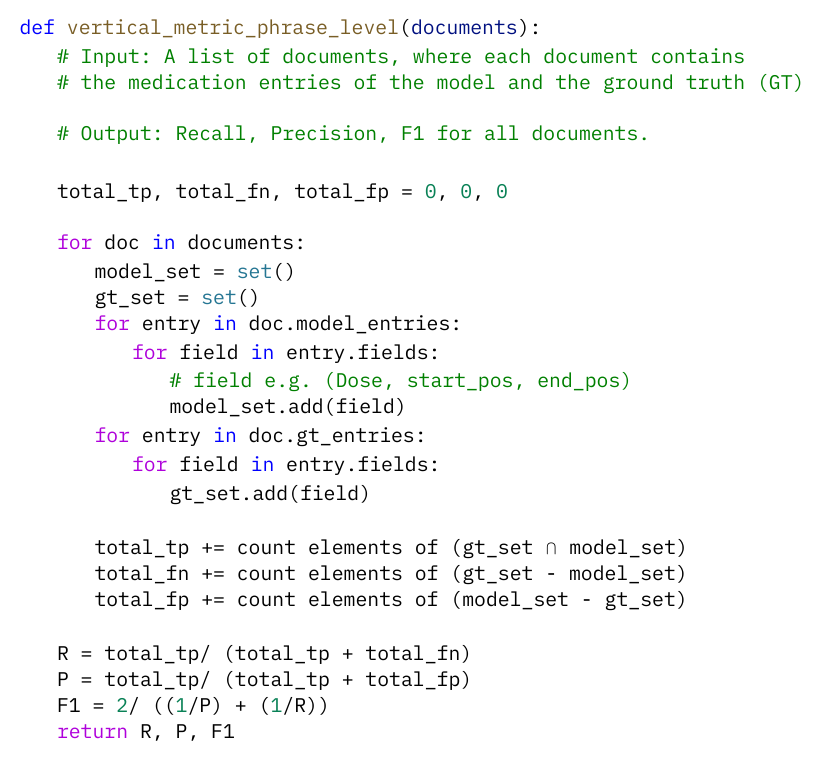} 
    \caption{Pseudocode for Vertical metrics.}
    \label{fig:vertical_metrics}
\end{figure}

\begin{figure}[h!]
    \centering
    \includegraphics[width=0.5\textwidth]{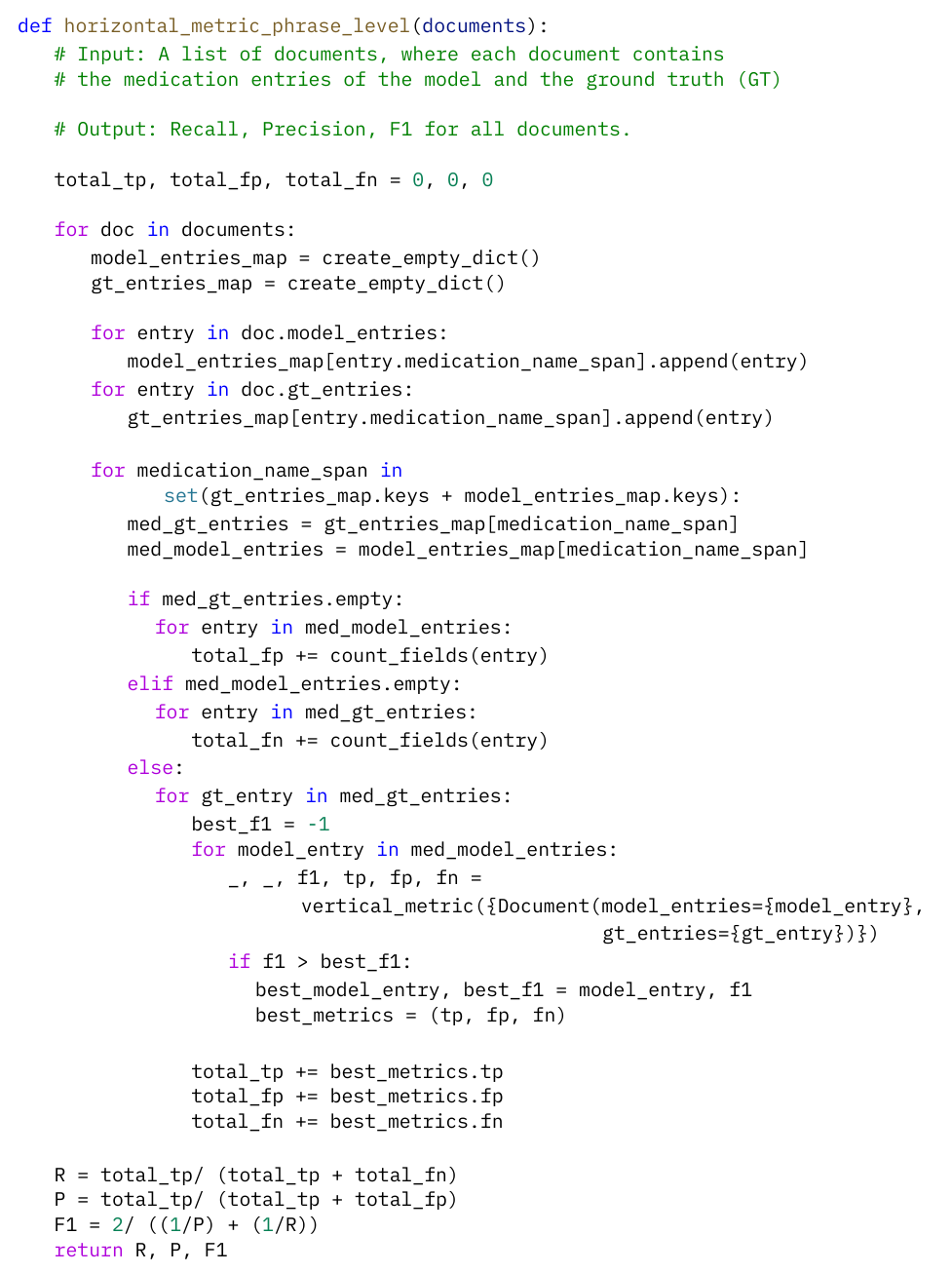} 
    \caption{Pseudocode for Horizontal metrics.}
    \label{fig:horizontal_metrics}
\end{figure}

Figures~\ref{fig:vertical_metrics} and \ref{fig:horizontal_metrics} present pseudocode for the Vertical metrics and Horizontal metrics, respectively. Both are based on the metrics defined by ~\citep{uzuner2010extracting}. As described in Section \ref{sec:quality_metrics}, \emph{Vertical metrics} evaluate NER with respect to fields, while \emph{horizontal metrics} also evaluate RE between fields and the medication entry.  

\section{More Details on Statistical Analysis}
\label{app:stats}

Here we provide more details on the statistical analysis conducted in \Cref{sec:res}.

\subsection{Statistical Significance Test}
To compare the performance metrics (precision, recall and F1 score) of various labeling methods, we use the approximated randomization test in \citep{yeh2000more}, which is more powerful and precise than the method implemented in \textit{i2b2}. For two different labeling methods $A$ and $B$, denote their output sets as $O_A$ and $O_B$ and the performance metrics as $f_A$ and $f_B$. Then $\Delta = f_A - f_B$ is the observed performance difference, $Q = O_A \cap O_B$ is the set of common outputs shared by both methods, and
$P_A = O_A\setminus O_B$ and $P_B = O_B\setminus O_A$ are the sets of unique outputs only generated from either method $A$ or method $B$. When the two methods are similar (null hypothesis), the outputs in $P_A$ and $P_B$ should be exchangeable. The implementation details of our approximated randomization test are as follows:
\begin{enumerate}
\itemsep0em 
\item Gather unique outputs generated from either $A$ or $B$, i.e., $P=P_A \cup P_B$.
\item Randomly assign each output in $P$ to new sets $P^\prime_A$ or $P^\prime_B$ with equal probability, and create the set of psedo-outputs for $A$ and $B$ 
$$O^\prime_A = P^\prime_A \cup Q, \: O^\prime_B = P^\prime_B \cup Q.$$
\item Compute performance metrics on $O^\prime_A$ and $O^\prime_B$, and calculate their difference $\Delta^\prime$.
\item Repeat Steps 2-3 for $n=1,000$ times, and generate a sampling distribution of $\Delta^\prime$.
\end{enumerate}
If the observed metric difference $\Delta$ is above the $95\%$ percentile of the sampling distribution, we can claim that method $A$ performs significantly better than $B$ with metric difference $\Delta$.

\subsection{Time cost and types of refinement}
\label{app:typesrefinment}
\begin{figure}[h!]
    \centering
    \includegraphics[width=0.45\textwidth]{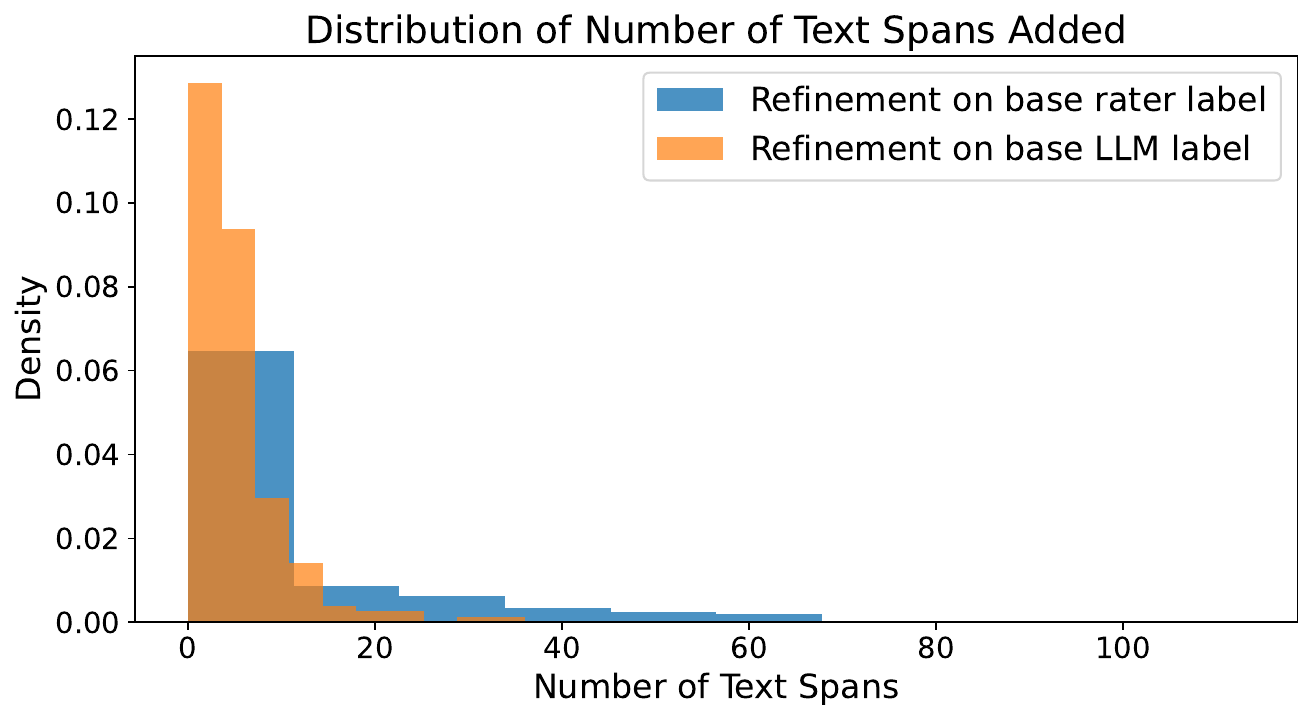} 
    \caption{Distribution of additions in the refinement process, LLM vs. rater base generated labels.}
    \label{fig:dist_corrections_add}
\end{figure}

\begin{figure}[h!]
    \centering
    \includegraphics[width=0.45\textwidth]{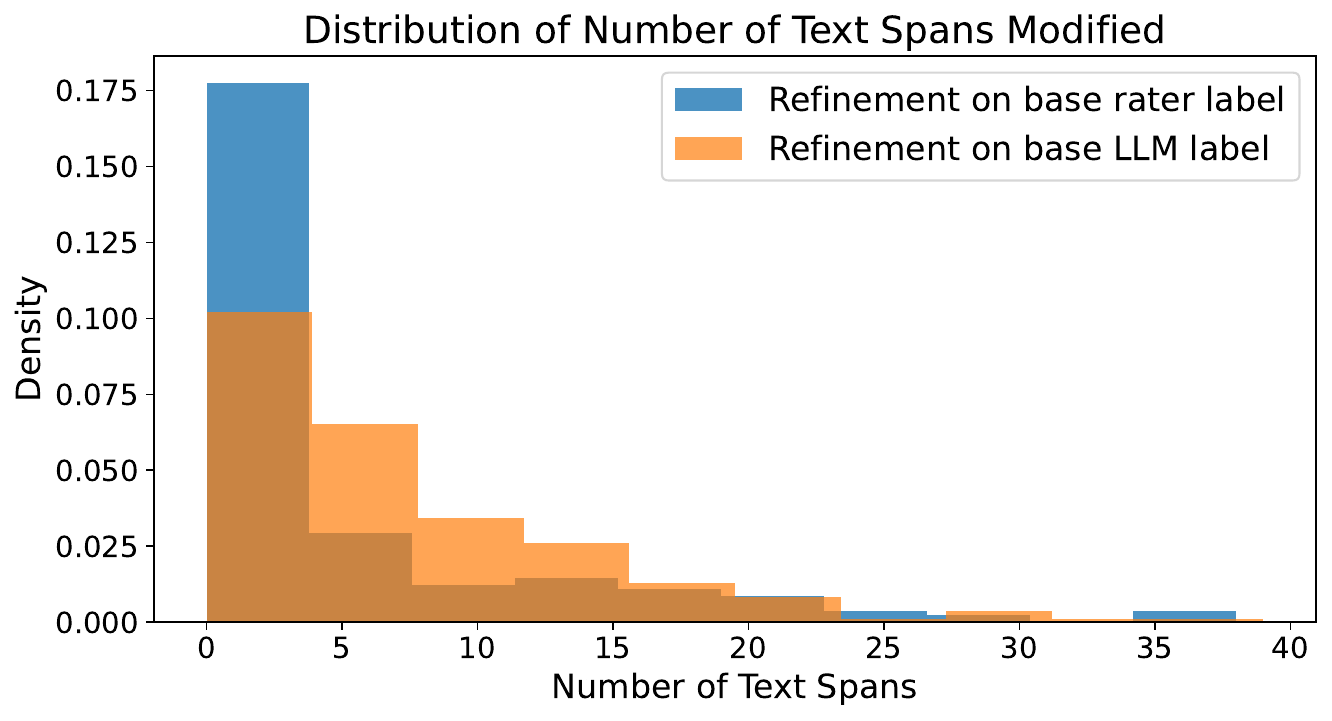} 
    \caption{Distribution of modifications in the refinement process, LLM vs. rater base generated labels.}
    \label{fig:dist_corrections_mod}
\end{figure}

\begin{figure}[h!]
    \centering
    \includegraphics[width=0.45\textwidth]{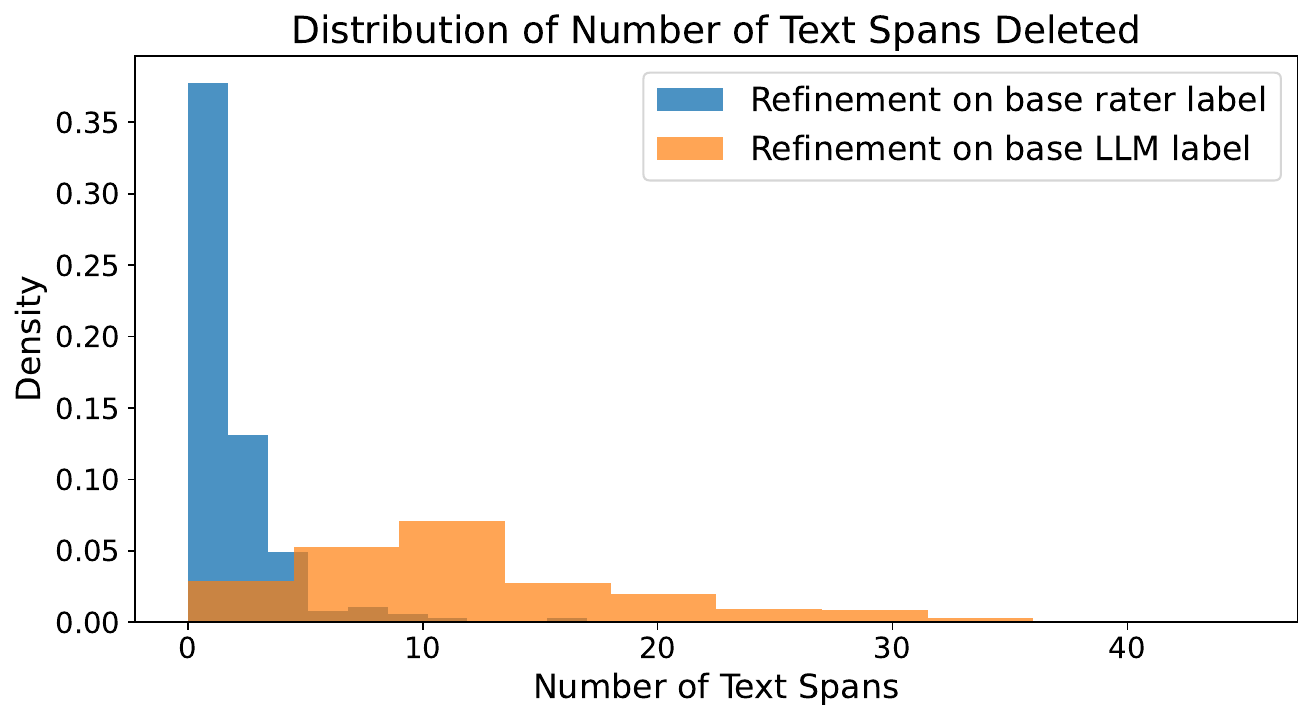}
    \caption{Distribution of deletions in the refinement process, LLM vs. rater base generated labels.}
    \label{fig:dist_corrections_del}
\end{figure}

We conducted further analysis to investigate what type of errors in pre-labels / what type of corrections has greatest impact on time cost. Focusing on text spans only, there are three types of corrections a human refiner can do: addition (adding a missing text span), modification (modify an existing text span) and deletion (delete a spurious text span). First, we look into the distribution of these three types of corrections for refinement based on human pre-labels and refinement based on LLM pre-labels Figures ~\ref{fig:dist_corrections_add}, \ref{fig:dist_corrections_mod}, and \ref{fig:dist_corrections_del}. Refinement on LLM pre-labels usually requires significant more deletions and modifications, while the human + refinement method involves more additions (all p-values $<0.5$). By fitting a regression model on time cost using the number of corrections of each type as predictors, the amount of time needed for adding a missing text span and modify an existing text span are similar, while it is much faster to delete a text span; see regression results in Table \ref{tab:time_cost_regression}.

\begin{table}[h]
    \label{app.labeltypetime}
    \centering
    \begin{tabular}{l|ccc}
    \toprule
        Variable & Coefficient & Std. Error & z score \\
        \midrule
        Intercept & 4.327** & 0.735 & 5.888  \\
        \# Modified X & 0.351** & 0.076 & 4.618 \\
        \# Added & 0.288** & 0.031 & 9.158 \\
        \# Deleted & -0.013 & 0.235 & -0.056 \\
        \bottomrule
    \end{tabular}
    \caption{Linear mixed effect model regression results: time cost vs. type of corrections. ** denotes $p<0.01$.}
    \label{tab:time_cost_regression}
\end{table}

\begin{table*}[!ht]
\centering
\begin{tabular}{lrrrrr}
\toprule
       \textbf{Method} &  \textbf{Medication} &  \textbf{Dose} &  \textbf{Frequency} &  \textbf{Mode} &  \textbf{Duration} \\
\midrule
          Base LLM & 0.784 & 0.883 & 0.787 & 0.848 & 0.420 \\
        Base Rater & 0.776 & 0.903 & 0.883 & 0.907 & 0.384 \\
         Refined Rater &  0.908 &  0.927 &  0.914 &  0.952 &  0.466 \\
         Refined LLM &  0.907 &  0.941 &  0.890 &  0.947 &  0.540 \\
\bottomrule
\end{tabular}
\caption{F1 scores for different types of labels across each medication attribute. The attributes include medication name, dosage, frequency, mode, and duration.}
\label{tab:full_notes_vertical_per_attribute}
\end{table*}

\begin{table*}[!ht]
\centering
  \begin{tabular}{l|ccc|ccc|cc}
    \multicolumn{1}{c|}{\textbf{Label Type}} &
      \multicolumn{3}{c|}{\textbf{Vertical}} &
      \multicolumn{3}{c|}{\textbf{Horizontal}} \\
    & Recall & Precision & F1 & Recall & Precision & F1  \\
    \midrule
    Base Rater &  0.820429 &     0.924383 & 0.869309 &  0.723582 &     0.756489 & 0.739669 \\
     Base LLM &  0.874945 &     0.769727 & 0.818970 &  0.800494 &     0.678708 & 0.734587 \\
Refined Rater &  0.922830 &     0.925182 & 0.924005 &  0.873006 &     0.869832 & 0.871416 \\
  Refined LLM &  0.939259 &     0.900770 & 0.919612 &  0.886027 &     0.843396 & 0.864186 \\
    \bottomrule
  \end{tabular}
  \caption{\emph{Token-level} labeling Quality and Time Metrics for the Entire Test Set.}
 \label{tab:overall_metrics_token}
\end{table*}

\section{Additional Results}
\label{app:additionalres}

For brevity, we provide here additional results that support the main findings in \Cref{sec:res}.

\subsection{Supplemental Results showing performance across each field type}

In \Cref{tab:full_notes_vertical_per_attribute}, we compare all of the tested approaches in terms of $F1$ scores for each entity type.

\subsection{Supplemental Results showing token-level metrics} \label{app:token_level_evaluation}

\Cref{tab:overall_metrics_token} presents token-level labeling quality in terms of horizontal and and vertical metrics on the test set. Results are similar to the ones of phrase-level metrics.


\subsection{Time Cost Evaluations}
\label{app:time_cost_raters}

\begin{figure}[!htbp]
\centering
\includegraphics[width=\linewidth]{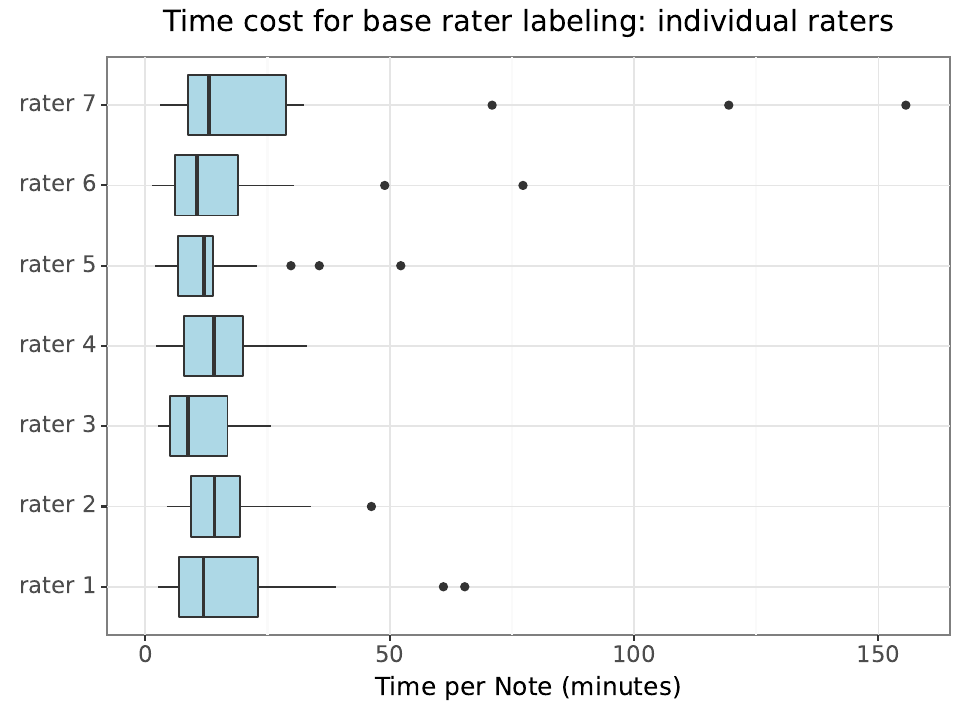}
\caption{Base labeling time cost of individual raters. Measurements are in minutes per note.}
\label{fig:time_cost_per_rater}
\end{figure}

\begin{figure}[!htbp]
\centering
\includegraphics[width=\linewidth]{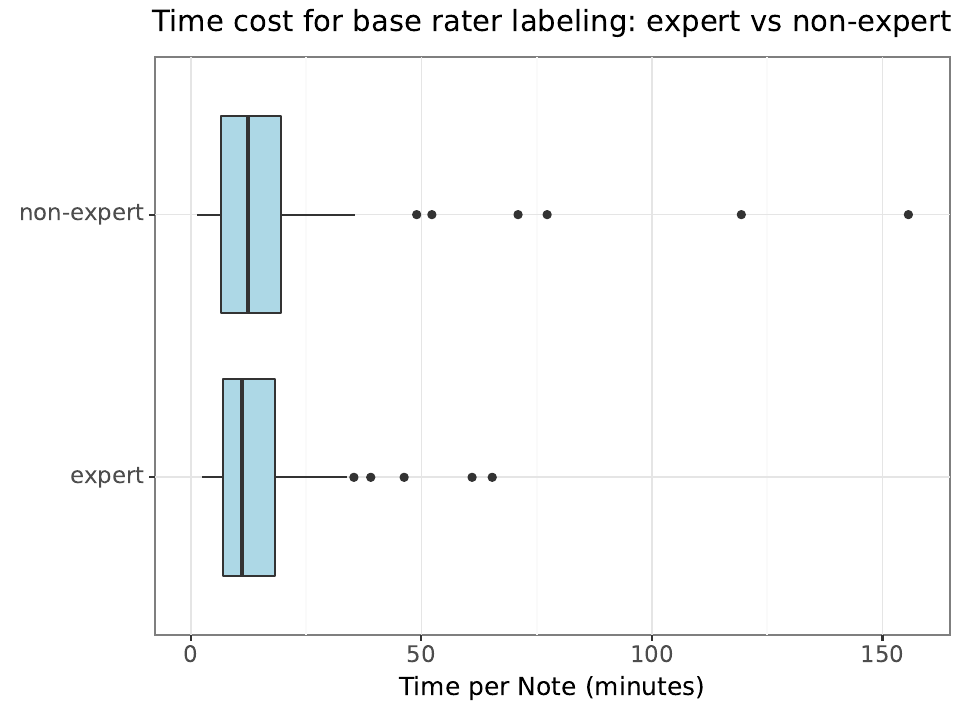}
\caption{Base labeling time cost: expert raters vs non-expert raters. Measurements are in minutes per note.}
\label{fig:time_cost_expert_nonexpert}
\end{figure}

In \Cref{fig:time_cost_per_rater} and \Cref{fig:time_cost_expert_nonexpert}, we show the time cost for individual raters and based on their level of expertise, respectively.

\onecolumn
\section{ Example Prompts}
\label{app:promptexample}

Note: The following examples are synthetic and inspired by, but not directly taken from, the \emph{i2b2} dataset. Only a single question-answer example is provided in each prompt for the sake of brevity.

\subsection{IOB-Token Prompt Example}

\begin{lstlisting}[breaklines=true, basicstyle=\ttfamily\small, frame=single]
description: "You are a helpful medical knowledge assistant for Beginning-Inside-Outside (BIO) tagging of medications' information. Your task is to create Medical Named Entity Recognition (NER) annotations using the Beginning-Inside-Outside (BIO) format. Label the given sentence (question) using BIO tags from these classes ONLY: MEDICATION (medication name), DOSE, FREQUENCY, DURATION, REASON, MODE. For tokens not belonging to these classes, tag as O-tag. For each token you MUST start a new line. Punctuations should be skipped in tokens. Each line must include: (1) word or token, (2) BIO tag, and (3) entity number for B-tags and I-tags ONLY. Use '<None>' for O-tags. Maintain the correct format. Example: 'token', BIO-tag, 'entity_1'." 

examples: {
  question: "'CURRENT MEDICATIONS: Ibuprofen as needed, and diclofenac for one month as needed, for abdominal discomfort.'"
  answer:
    "'CURRENT', O, '<None>'\n"
    "'MEDICATIONS', O, '<None>'\n"
    "'Ibuprofen', B-MEDICATION, 'entity_1'\n"
    "'as', B-FREQUENCY, 'entity_1'\n"
    "'needed', I-FREQUENCY, 'entity_1'\n"
    "'and', O, '<None>'\n"
    "'diclofenac', B-MEDICATION, 'entity_2'\n"
    "'for', B-DURATION, 'entity_2'\n"
    "'one', I-DURATION, 'entity_2'\n"
    "'month', I-DURATION, 'entity_2'\n"
    "'as', I-DURATION, 'entity_2'\n"
    "'needed', I-DURATION, 'entity_2'\n"
    "'for', O, '<None>'\n"
    "'abdominal', B-REASON, 'entity_1|entity_2'\n"
    "'discomfort', I-REASON, 'entity_1|entity_2'\n"
}
\end{lstlisting}

\newpage
\subsection{Direct-Chunk Prompt Example}

\begin{lstlisting}[breaklines=true, basicstyle=\ttfamily\small, frame=single]
description:
  "You are a medical assistant with expertise in document processing.\n"
  "\n"
  "- Your task is to tag groups of entities related to medications. (Some groups may have overlap as two medications can be used for the same reason.)\n"
  "- Each medication is a SPECIFIC reference to the drug's name used by the patient. This should not contain other data already within a group.\n"
  "- The mode of administration, such as 'oral' or 'intravenous', should be tagged SEPARATELY from the medication's name. \n"
  "- Text should contain the EXACT characters in the raw input, including punctuation, and typos.\n"
  "- Use YAML as the output format."

examples: {
  question: "'CURRENT MEDICATIONS: Ibuprofen as needed, and diclofenac for one month as needed, for abdominal discomfort.'"
  answer:
    "```yaml\n"
    "entities:\n"
    "  - group: 1\n"
    "    MEDICATION: \n"
    "      text: Ibuprofen\n"
    "      start_pos: 20\n"
    "      end_pos: 29\n"
    "    DOSE: \n"
    "      text: ''\n"
    "      start_pos: \n"
    "      end_pos: \n"
    "    FREQUENCY: \n"
    "      text: as needed\n"
    "      start_pos: 30\n"
    "      end_pos: 39\n"
    "    DURATION: \n"
    "      text: ''\n"
    "      start_pos: \n"
    "      end_pos: \n"
    "    REASON: \n"
    "      text: abdominal discomfort\n"
    "      start_pos: 81\n"
    "      end_pos: 100\n"
    "    MODE: \n"
    "      text: ''\n"
    "      start_pos: \n"
    "      end_pos: \n"
    "  - group: 2\n"
    "    MEDICATION: \n"
    "      text: diclofenac\n"
    "      start_pos: 45\n"
    "      end_pos: 55\n"
    "    DOSE: \n"
    "      text: ''\n"
    "      start_pos: \n"
    "      end_pos: \n"
    "    FREQUENCY: \n"
    "      text: as needed\n"
    "      start_pos: 68\n"
    "      end_pos: 77\n"
    "    DURATION: \n"
    "      text: for one month\n"
    "      start_pos: 56\n"
    "      end_pos: 68\n"
    "    REASON: \n"
    "      text: abdominal discomfort\n"
    "      start_pos: 81\n"
    "      end_pos: 100\n"
    "    MODE: \n"
    "      text: ''\n"
    "      start_pos: \n"
    "      end_pos:\n"
    "```\n"
}
\end{lstlisting}

\twocolumn
\let\clearpage\relax
\end{document}